\documentclass[11pt,a4paper]{article}
\usepackage[hyperref]{emnlp2020}
\usepackage{times}
\usepackage{latexsym}
\usepackage{graphicx} 
\usepackage{fixltx2e}

\usepackage{microtype}

\aclfinalcopy 

\title{May I Ask Who's Calling? Named Entity Recognition on Call Center Transcripts for Privacy Law Compliance}

\author{Micaela Kaplan \\
  Brandeis University / Waltham, MA \\
  CallMiner Inc / Waltham, MA \\
  \texttt{micaela@brandeis.edu} \\}

\date{}

\begin{document}
\maketitle
\begin{abstract}
We investigate using Named Entity Recognition on a new type of user-generated text: a call center conversation. These conversations combine problems from spontaneous speech with problems novel to conversational Automated Speech Recognition, including incorrect recognition, alongside other common problems from noisy user-generated text. Using our own corpus with new annotations, training custom contextual string embeddings, and applying a BiLSTM-CRF, we match state-of-the-art results on our novel task.
\end{abstract}

\section{Introduction}
When a call center says ``a call may be recorded", they are often collecting a transcript. These transcripts are the output of speech recognition systems, and while they are redacted for Payment Card Industry (PCI) compliance, they often contain other information about the caller such as their name and internal ID number. This data can be helpful for quality assurance and future customer care. New privacy laws, such as the General Data Protection Regulation (GDPR) in the EU, define rules and regulations for everything from how data is collected and stored to the rights of a person to retract their consent to the use of their data \citep{gdpr.eu_2019}. In the face of these new laws, it is important to be able to identify non-public personal information and personally identifiable information (NPI/PII) in call transcripts in order to comply with regulations without compromising the data these companies rely on.

We use Named Entity Recognition (NER) to find instances of NPI/PII, remove them from a transcript, and replace them with a tag identifying which type of information was removed. For example, a transcript containing ``This is john doe reference number 12345" would become ``This is [NAME] reference number [NUMBER]". This problem is unique in a call center for a few reasons. Firstly, call transcripts are organic human conversations and present many of the common problems of user-generated data, including false starts, incomplete sentences, and novel words. Secondly, the text provided in a transcript is the output of an Automatic Speech Recognition (ASR) system, which is prone to error as described in Section \ref{annotation}. While modern ASR systems are reliable, our input audio is from phone calls, which are usually very low-quality and often contain a lot of background noise. This low-quality audio results in poor ASR, which then outputs sentences that may not be grammatical. This makes it difficult to understand the semantics of the call or to pick up on many of the features that are critical to most NER systems such as context or part of speech. Additionally, production-level call transcripts, or those that are used by Quality Assurance agents and data scientists, are missing capital letters, numeric digits, and accurate punctuation, which are features that are crucial to the classic approaches to NER. Moreover, traditional NER systems use labels for proper nouns, like people's names, but have no way to handle emails, spellings, or addresses, making bootstrapping from pretrained NER models impossible.

In this paper, we apply the current state-of-the-art neural architecture for sequence labeling, a BiLSTM-CRF, to our novel call center transcripts in search of NPI and PII as identified by a human. We match state-of-the-art performance for standard datasets on our novel problem by using our model in conjunction with annotated data and custom contextual string embeddings.

\section{Previous Work}\label{Previous Work}

NER became popular in the NLP community at the Message Understanding Conferences (MUCs) during the 1990s \cite{HIRSCHMAN1998281}. In 2003, the CoNLL2003 shared task focused on language independent NER and popularized feature based systems \cite{tjong-kim-sang-de-meulder-2003-introduction}. The OntoNotes corpus, released in 2006, has also been fundamental to NER research \citep{hovy2006ontonotes}. 

After CoNLL, the highest performing models were based on a CRF \cite{lafferty-et-al} which requires the manual generation of features. More recently, research has used neural techniques to generate these features. \citet{huang2015bidirectional} found great success using Bidirectional Long Short Term Memory models with a CRF layer (BiLSTM-CRF) on both the CoNLL2000 and CoNLL2003 shared task datasets. \citet{ma2016endtoend} used a BiLSTM-CNN-CRF to do NER on the CoNLL2003 dataset, producing state-of-the-art results. Similarly, \citet{chiu2015named} used a BiLSTM-CNN, with features from word embeddings and the lexicon, which produced very similar results. \citet{ghaddar-langlais-2018-robust} used embeddings for the words and for entity types to create a more robust model. Flair, proposed by \citet{akbik2018coling}, set the current state of the art by using character based embeddings, and built on this with their pooling approach in 2019 \cite{akbik2019naacl}. Crossweigh, a framework introduced by \citet{wang-etal-2019-crossweigh}, makes use of Flair embeddings to clean mishandled annotations.

In 2006, \citeauthor{sudoh2006incorporating} used the word confidence scores from ASR systems as a feature for NER on the recordings of Japanese newspaper articles. In 2018, \citet{ghannay2018endtoend} conducted a similar experiment on French radio and TV audio. Unlike our data, neither of these tasks used spontaneous conversation. Additionally, the audio was probably recording-studio quality, making ASR a reliable task. 

\subsection{Conversations are Different: The Twitter Analogy}
All of the previous work discussed was run on datasets primarily comprised of newswire data \citep{li2018survey}. Typically, newswire follows the conventions of normal text, but call center transcripts have none of these conventions guaranteed and often explicitly lack them entirely. This is a problem for the traditional approaches to NER. Our low-quality audio adds to the difficulty of this task.

The closest approximation of this problem in the previous research is on Twitter data. Tweets, like transcripts, are generated by users and may not follow the grammar, spelling, or formatting rules that newswire is so careful to maintain. In 2011, \citet{liu-etal-2011-recognizing} used a K-nearest neighbors model combined with a CRF to begin tackling this problem. As part of the 2017 Workshop on Noisy User-generated Text (W-NUT) shared task, \citet{Aguilar_2017} obtained a first place ranking using a model that combined a multi-step neural net with a CRF output layer. \citet{akbik2019naacl} also tested their pooled contextualized string embeddings on this data and found success. We use this previous work on tweets to inform our model creation for the call center space. 

\section{Data}\label{data}
Our dataset consists of 7,953 training, 500 validation, and 534 test samples. Each sample is a complete speaker turn from a call taken by a call center that deals with debt collection. For our purposes, a speaker turn is defined as the complete transcription of one speaker before another speaker starts talking, as illustrated in Figure \ref{fig:turns}. The training set is a random sample of turns from 4 months of call transcripts from the client, but was manually curated to contain examples of NPI/PII to compensate for its relatively rarity in call center conversation. The transcripts were made using a proprietary speech recognition system, which is set to provide all lowercase transcripts and omits punctuation and numeric digits. We used spaCy\footnote{\url{https://spaCy.io/}} to convert each turn to a document that starts with a capital letter and ends with a period. This is due to the default configurations of spaCy-- in order to make use of entities, we needed to add in a Sentencizer module, which defaults to this capital letter and period set up.

\begin{figure}
\vspace*{-1.5em}
\textbf{Speaker 1:} Thank you for calling our company how may i help you today.

\textbf{Speaker 2:} Id like to pay my bill.

\caption{An example of turns of a conversation, where each person's line in the dialogue represents their turn. This output matches the format of our data described in Section \ref{data}.}
\label{fig:turns}
\vspace*{-1.5em}
\end{figure}

\subsection{Data Annotation}\label{annotation}
\begin{table}
    \vspace{-1.5em}
    \begin{tabular}{|p{2cm}|p{5cm}|}
       \hline
       Entity Type & Description \\
       \hline
       NUMBERS & A sequence of numbers relating to a customer's information (e.g. phone numbers or internal ID number)\\
       NAME & First and last name of a customer or agent\\
       COMPANY & The name of a company\\
       ADDRESS & A complete address, including city, state, and zip code\\
       EMAIL & Any email address\\
       SPELLING & Language that clarifies the spelling of a word, 
       (e.g. ``a as in apple")\\
    \hline
    \end{tabular}
    \caption{A brief description of our annotation schema.}
    \label{table:annotation}
    \vspace{-1em}
\end{table}
We created a schema to annotate the training and validation data for a variety of different categories of NPI/PII as described in Table \ref{table:annotation}. Initial annotation was done with Doccano\footnote{\url{http://doccano.herokuapp.com/}}. These annotations were converted to entities in the text with spaCy. The annotators were trained in NPI/PII recognition, and were instructed to lean towards a greater level of caution in ambiguous cases. This ambiguity was often caused by misrecognitions from the language model in the ASR system being used. With no audio to help the annotator, it wasn't always clear when ``I need oak leaves" was supposed to be ``Annie Oakley". The reverse problem was also true. ``Brilliant and wendy jeff to process the refund" appears to be a full name, but is actually a misrecognition for ``Brilliant and when did you want to process the refund". Emails also proved difficult, because misrecognitions made it difficult for annotators to discern exactly what words belonged in the email address. Another difficulty for annotation was that all of our data had been pre-redacted for PCI compliance, which requires the redaction of number strings relating to credit card numbers, birth dates, and social security numbers. This redaction occurs before any transcript can be released to a client or researcher. To minimize false negatives, PCI redaction frequently redacts numbers that are NPI/PII such as in an internal customer ID number or a phone number. Since the NUMBERS label was intended to catch these NPI/PII related numbers, we used context to include this [redacted] tag as part of a numbers sequence when possible. No steps to clean the transcripts were taken at any point. The naturally occurring noise in our data is critical to our use case and was left for the model to interpret. 

Due to limitations with spaCy and the known complexity of nested entities, we opted to allow only one annotation per word in our dataset. This means that ``c a t as in team at gmail dot com" could only be labeled either as SPELLING[0:6] EMAIL[6:] or as EMAIL[0:] with indices that correspond to the location of the word in the text and are exclusive. This ultimately explains the much lower number of SPELLING entities in the dataset as compared to other entities, because they are often contained a part of EMAIL or ADDRESS. This will influence our analysis in Section \ref{discussion}.

\section{Model Design}
\begin{figure}
    \vspace{-1.5em}
    \centering
    \resizebox{\columnwidth}{!}{
    \includegraphics{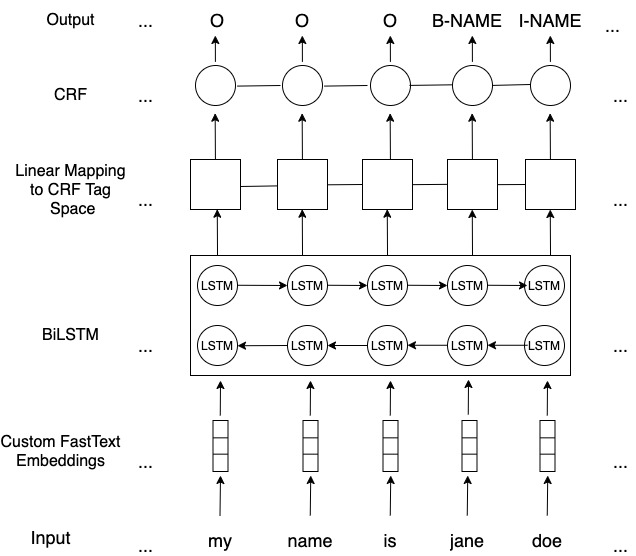}
    }
    \caption{A schematic of our BiLSTM-CRF model. The text of each turn is passed to a word embedding layer which is followed by a BiLSTM layer, and then a linear layer that maps the word BiLSTM output into tag space. Finally, the CRF layer produces an output sequence.}
    \label{fig:model}
    \vspace{-1.5em}
\end{figure}

We implemented a standard BiLSTM-CRF model in PyTorch. The basic model implementation is adapted from a GitHub  repository\footnote{\url{https://github.com/mtreviso/linear-chain-crf}}. We wrote our own main.py to better allow for our spaCy preprocessing, and we also adapted the code to handle batch processing of data. After this preprocessing, we trained the model with the training set and used the validation set for any model tuning. All reported numbers are on the test set and occur after all tuning is completed.  A visualization of our model is found in Figure \ref{fig:model}.

\section{Experiments} \label{experiments}
\subsection{Basic Hyperparameter Tuning}\label{og}
We used a grid search algorithm to maximize the performance of the model. The word embedding layer uses FastText embeddings trained on the client's call transcripts. We find that this helps mitigate the impacts of poor ASR in other aspects of our research, and investigate this further in Sections \ref{word embeddings} and \ref{flair}. The grid search contained the following parameters: epochs (a sampled distribution between 5 and 50), the size of a dropout layer (between 0 and .5, with .1 intervals of search), the number of hidden layers (between 5 and 20 in increments of 5), and the encoding type used in the output of the CRF (BIO, BILOU, IO). The other hyperparameters in our model were learning rate .001, batch size 1, 30 nodes in each fully connected layer, and the inclusion of bias in each layer. The experiments were run in parallel using Python's multiprocessing package on a virtual machine with 16 CPUs and 128GB of memory. Each experimental configuration ran on a time scale of a few hours, relative to the configurations of the hyperparameters being used.

To better understand the performance of our model on the test set, we broke down the precision, recall, and F1 measurements by entity type. Table \ref{table:embeddings} shows these results for the best model configuration under the columns labeled ``Custom". This model used 46 epochs, a dropout rate of .2, 5 hidden layers, and a BIO encoding.

\subsection{Training Word Embeddings}\label{word embeddings}
While much of the previous research has fine-tuned existing word embeddings, the task of compensating for misrecognition seemed less straightforward than domain adaptation. We lessen the impact of the misrecognitions described in Section \ref{annotation} by understanding that frequent misrecognitions appear in contexts similar to the intended word. For example, ``why you're" is often misrecognized as ``choir" which would have a totally out of context vector from a pretrained model in this data set. A custom model gives ``choir" a vector that is more similar to ``why" than to ``chorus". \citet{embed} showed the importance of domain specific word embeddings when using ASR data.

We ran our best performing model configuration with the 300 dimensional GloVe 6b word embeddings \footnote{\url{https://nlp.stanford.edu/projects/glove/}}. Our embeddings, in contrast, are trained on approximately 216 million words, making them substantially smaller than other state-of-the-art embeddings used today. The results from the best epoch of this model (16) are shown in Table \ref{table:embeddings}.

\begin{table}
    \vspace{-1.5em}
    \resizebox{\columnwidth}{!}{
    \begin{tabular}{|c|c c|c c|c c|}
    \hline
     Entity Type & \multicolumn{2}{c|}{Precision} & \multicolumn{2}{c|}{Recall} & \multicolumn{2}{c|}{F1} \\
     \hline
      & Custom & GloVe & Custom & GloVe & Custom & GloVe\\
     \cline{2-7}
     O & 89.8 & 84.2 & 81.7 & 76.6 & \textbf{85.6} & 80.2\\
     NUMBERS & 95.6 & 88.7 & 85.4 & 82.9 & \textbf{90.1} & 85.7\\
     NAME & 89.6 & 92.1 & 91.1 & 88.7 & 90.3 & 90.3\\
     COMPANY &98.8&99.5&72,9&64.3&\textbf{83.9}&78.1\\
     ADDRESS &70.6&.3&75.0&18.7&\textbf{72.7}&23 \\
     EMAIL\footnotemark & 0 & 07.1& 0 &03.1& 0&\textbf{04.4}\\
     SPELLING &45.8&.34&52.4&40/5&\textbf{48.9}&37.0\\
     \hline
     Micro Average & 89.2&85.6 &79.6 &74.0& \textbf{84}.1&79.4\\
    \hline
    \end{tabular}
    }
    \caption{The performance by entity type of the BiLSTM-CRF model on the held out test set. This table compares the results of our custom embeddings model (``Custom") against the GloVe embeddings (``GloVe").}
    \label{table:embeddings}
    \vspace{-1.5em}
\end{table}

\footnotetext{Our custom model gets all 0s because many of its predicted EMAIL entities were off by a few words. We discuss this more in Section \ref{discussion}.}

\subsection{Using Flair}\label{flair}
In our previous experiments, we established the importance of using custom word embeddings to accurately account for the misrecognitions, false starts, and other kinds of noise present in call center conversation transcripts. In this experiment, we test the performance of Flair\footnote{\url{https://github.com/flairNLP/flair}} and its contextual string embeddings on our data. 

We begin by training custom contextual string embeddings for this dataset, based on the findings in our original experiments. For training, we use the same corpus as used in Section \ref{og}. To do this we follow the tutorial on the Flair GitHub page using their suggested hyperparameter settings as follows: hidden\_size: 1024, sequence\_length: 250, mini\_batch\_size: 100, and otherwise use the default parameters. We use the newline to indicate a document change, and list each turn as a separate document to provide consistency with the other experiments conducted in this paper. Given the size of our corpus for word embedding training, we found that our model's validation loss stabilized after epoch 4. We use the best version of model, as given by Flair, in all of our tests.

We conduct a number of experiments using Flair's SequenceTagger with default parameters and a hidden\_size of 256. We adapt the work done by \citet{akbik2018coling} and \citet{akbik2019naacl} to explore the impact of call center data on these state-of-the-art configurations.

\textbf{Flair} uses only the custom trained Flair embeddings.

\textbf{Flair\textsubscript{+ FastText}} uses the custom trained Flair embeddings and our custom trained FastText embeddings using Flair's StackedEmbeddings.

\textbf{Flair\textsubscript{mean pooling}} uses only the custom trained Flair embeddings within Flair's PooledFlairEmbedding. We use mean pooling due to the results of \citet{akbik2019naacl} on the WNUT-17 shared task.

\textbf{Flair\textsubscript{mean pooling + FastText}} uses the PooledFlairEmbeddings with mean pooling and the custom trained FastText embeddings using Flair's StackedEmbeddings.

These results are shown in Table \ref{table:flair}.
\begin{table}[t]
    \vspace{-1.5em}
    \resizebox{\columnwidth}{!}{
    \begin{tabular}{|c|c|c|c|c|}
    \hline
    Entity & Flair & Flair\textsubscript{+ FastText} & Flair\textsubscript{mean pooling} & Flair\textsubscript{mean pooling + FastText}\\
    \hline
    O& 98.3 & 98.5 & 98.2 & 98.5\\
    NUMBERS & 83.1 & \textbf{87.9} & 87.7 & 86.2\\
    COMPANY & \textbf{81.1} & 80.7 & 80.7 & 80.3\\
    ADDRESS & 87.5 & \textbf{94.1} & 61.5 & \textbf{94.1}\\
    EMAIL & 58.8 & 50.0 & \textbf{73.3} & 66.7 \\
    SPELLING & 55.0 & 57.1 & 55.8 & \textbf{57.9}\\
    Micro Average & 97.5 & 97.7 & 97.3 & 97.7\\
    \hline
    \end{tabular}
    }
    \caption{The F1 scores on the test set for each entity type for each Flair embedding experiment.}
    \vspace{-1.5em}
    \label{table:flair}
\end{table}

\section{Discussion}\label{discussion}
Table \ref{table:embeddings} shows that in all cases except for EMAIL, it is beneficial to use our custom embeddings over GloVe embeddings. We explain this in the next paragraph. The Flair embeddings show a large improvement over the other word embedding varieties however in our circumstance all four varieties of Flair models have nearly identical Micro Average F1s. The best performing Flair models are those that use both the custom contextualized string embeddings and the custom FastText embeddings.

Across all of the models in this paper, EMAIL and SPELLING consistently performed worse than other categories. We believe this is due to the overlap in their occurrences as well as the variability in their appearance. In many cases the custom embeddings model identified parts of an email correctly but attributed certain aspects, like a name, as NAME followed by EMAIL instead of including them together as EMAIL. SPELLING often appears within an EMAIL entity, such as in ``c as in cat a t at gmail dot com". Due to the limitations discussed in Section \ref{data}, this leads to a limited occurrence of the SPELLING entity in our training data, and many EMAIL and ADDRESS entities that contain examples SPELLING. All models, especially the custom embeddings model, frequently misidentified EMAIL as SPELLING and vice versa. Additionally, our test data contained a number of turns that consisted of only SPELLING on its own, which was poorly represented in training. The Flair\textsubscript{mean pooling} model outperforms the other models in EMAIL by a large margin.

The results shown in Table \ref{table:flair} highlight other interesting notes about our data. The NUMBERS category contains many strings that appear consistently in the text. Not only are there a finite number of NUMBER words in our corpus (those numeric words along with many instances of ``[redacted]"), but the NUMBERS of interest in our dataset, such as account numbers, appear in very similar contexts and do not often get misrecognized. The COMPANY entity performs well for a similar reason. When the model was able to identify the company name correctly, it was usually in one of the very common misrecognition forms and in a known context, which furthers our claim that dataset specific embeddings give an important boost over pretrained embeddings. Where the models failed here can likely be attributed to training data. Since the name of the company is a proper noun, it is not in most standard ASR language models, including the one we use. Thus, it is a frequent candidate for misrecognition, because the language model has higher probabilities assigned to grammatically correct phrases that have nothing to do with the name of the company. This leads to high variability in appearance, which means that not every possible version of the company name was present in our training set.

Interesting variability also occurred in ADDRESS entities.
With ADDRESS, both models that used Flair and FastText embeddings strongly outperformed the models that used Flair on its own, but standard Flair embeddings strongly outperformed the Pooled Flair embeddings. Neither version of the Flair only model identified addresses in which house numbers or zip codes were shown as ``[redacted]" but both models that utilized FastText had no issue with these examples. 

\section{Conclusion and Future Work}
By using a BiLSTM-CRF, in conjunction with custom-trained Flair embeddings, we  match current state-of-the-art NER  performance on our novel call center conversation dataset with unique entity types. We also reinforce the importance of training word embeddings that fully capture the nuances of the data being used for the task. While we cannot release any data for privacy reasons, we have shown that current state-of-the-art techniques successfully carry over to more non-traditional datasets and tasks. In the future, we'd like to assess the contribution of this model with the call transcripts from other industries. Additionally, we'd like to investigate the success of these strategies on other user-generated conversations, such as chats and emails. 

\section*{Acknowledgments}
Thanks to the anonymous reviewers for their invaluable feedback. Thanks to CallMiner Inc. and its research partners for providing all of the data as well as the use cases and funding. Thanks to Jamie Brandon for her help with model architecture design. Thanks to the whole CallMiner research team for their help and support throughout the process.
\bibliography{anthology,emnlp2020}
\bibliographystyle{acl_natbib}

\end{document}